\documentclass[conference]{IEEEtran}
\IEEEoverridecommandlockouts
\usepackage{xcolor,soul,framed} %,caption

\colorlet{shadecolor}{yellow}
\usepackage[pdftex]{graphicx}
\graphicspath{{../pdf/}{../jpeg/}}
\DeclareGraphicsExtensions{.pdf,.jpeg,.png}

\usepackage[cmex10]{amsmath}
\usepackage{array}
\usepackage{mdwmath}
\usepackage{mdwtab}
\usepackage{eqparbox}
\usepackage{url}

\usepackage{amssymb}
\graphicspath{ {./images/} }
\usepackage{orcidlink}
\usepackage{breqn}
\usepackage{physics}

\usepackage{amsmath}

\usepackage{tikz} %for checkmark

\usepackage{float}

\usepackage{breqn}
\usepackage{physics}
\usepackage{subcaption}

 \usepackage{adjustbox}

\usepackage{booktabs}  % For \Xhline and improved table lines
\usepackage{array}     % For better column formatting
\usepackage{multirow}  % Optional: For multi-row functionality, not used here
\usepackage{graphicx}  % Optional: If you need to scale tables or include images

\usepackage{blindtext}
\usepackage{xcolor}
\usepackage{hyperref}
\usepackage{graphicx}
\graphicspath{ {./images/} }
\usepackage{diagbox}
\usepackage{amsmath} 
\usepackage{mathtools}
\usepackage{lipsum}
\usepackage[nocompress,space]{cite}
\usepackage{amssymb}
\usepackage{pdfrender}
\usepackage{blindtext}
\usepackage{xcolor}
\usepackage{hyperref}
\usepackage{graphicx}
\graphicspath{ {./images/} }
\usepackage{diagbox}
\usepackage{amsmath}
\usepackage{mathtools}
\usepackage{lipsum}
\usepackage[nocompress,space]{cite}
\usepackage{pdfrender}
\usepackage[
  separate-uncertainty = true,
  multi-part-units = repeat
]{siunitx}
\usepackage{textcomp}
\usepackage{siunitx}
\usepackage{float}
\usepackage{caption}
\usepackage{adjustbox}
\usepackage{rotating}
\usepackage{blindtext}
\usepackage{xcolor}
\usepackage{hyperref}
\usepackage{graphicx}
\graphicspath{ {./images/} }
\usepackage{diagbox}
\usepackage{amsmath}
\usepackage{mathtools}
\usepackage{lipsum}
\usepackage[nocompress,space]{cite}
\usepackage{pdfrender}
\usepackage{soul,color}
\usepackage{orcidlink}

\usepackage{xcolor,soul,framed} %,caption

\colorlet{shadecolor}{yellow}
\usepackage[pdftex]{graphicx}
\graphicspath{{../pdf/}{../jpeg/}}
\DeclareGraphicsExtensions{.pdf,.jpeg,.png}

\usepackage[cmex10]{amsmath}
\usepackage{array}
\usepackage{mdwmath}
\usepackage{mdwtab}
\usepackage{eqparbox}
\usepackage{url}

\usepackage{amssymb}
\graphicspath{ {./images/} }
\usepackage{orcidlink}
\usepackage{breqn}
\usepackage{physics}

\usepackage{tikz} %for checkmark

\usepackage{float}

\usepackage{multirow}
\usepackage{diagbox}
\usepackage{graphicx}  % For adjusting table sizes

\usepackage[T1]{fontenc}

\usepackage{threeparttable}  % for table footnotes

\usepackage{pifont}% http://ctan.org/pkg/pifont

\usepackage{bbding}
\usepackage{wasysym}

\usepackage{cleveref}

\usepackage{makecell} %thicker or thinner \hline

\newif\ifshowmods
\showmodsfalse   % uncomment this line to create a clean doc

\ifshowmods
                \newcommand{\Ablue}[1]{\textcolor{blue}{#1}}

\else

                \newcommand{\Ablue}[1]{#1}

\fi

 \makeatletter
\newcommand*\bigcdot{\mathpalette\bigcdot@{.5}}
\newcommand*\bigcdot@[2]{\mathbin{\vcenter{\hbox{\scalebox{#2}{$\m@th#1\bullet$}}}}}
\makeatother

\begin{document}

\title{

A Kolmogorov-Arnold Network for Explainable Detection of Cyberattacks on EV Chargers

}

\author{\IEEEauthorblockN{Ahmad Mohammad Saber\IEEEauthorrefmark{1},
Max Mauro Dias Santos\IEEEauthorrefmark{3},
Mohammad Al Janaideh\IEEEauthorrefmark{1}\IEEEauthorrefmark{4},
Amr Youssef\IEEEauthorrefmark{6} and
Deepa~Kundur\IEEEauthorrefmark{1} 
\IEEEauthorblockA{\IEEEauthorrefmark{1}Department of Electrical and Computer Engineering, University of Toronto, Toronto, ON, Canada}
\IEEEauthorblockA{\IEEEauthorrefmark{3}Department of Electronics, Federal Technological University of Paraná, Ponta Grossa, PR, Brazil}
\IEEEauthorblockA{\IEEEauthorrefmark{4}Department of Mechanical Engineering,  University of Guelph, Guelph, ON, Canada}
\IEEEauthorblockA{\IEEEauthorrefmark{6}Concordia Institute for Information Systems Engineering, Montreal, QC, Canada}
}
}

\maketitle

\begin{abstract}

The increasing adoption of Electric Vehicles (EVs) and the expansion of charging infrastructure and their reliance on communication expose  Electric Vehicle Supply Equipment (EVSE) to  cyberattacks. This paper presents a novel Kolmogorov-Arnold Network (KAN)-based framework for detecting cyberattacks on EV chargers using only power consumption measurements. Leveraging the KAN's capability to model nonlinear, high-dimensional functions and its inherently interpretable architecture, the framework effectively differentiates between normal and malicious charging scenarios. The model is trained offline on a comprehensive dataset containing over 100,000 cyberattack cases generated through an experimental setup. Once trained, the KAN model can be deployed within individual chargers for real-time detection of abnormal charging behaviors indicative of cyberattacks.
Our results demonstrate that the proposed KAN-based approach can accurately detect cyberattacks on EV chargers with Precision and F1-score of 99\% and 92\%, respectively, outperforming existing detection methods. Additionally, the proposed KANs's enable the extraction of mathematical formulas representing KAN's detection decisions, addressing interpretability, a key challenge in deep learning-based cybersecurity frameworks. This work marks a significant step toward building secure and explainable EV charging infrastructure.

\end{abstract}
\begin{IEEEkeywords}
Electric vehicle charging, Kolmogorov-Arnold network, cyber-physical security, machine learning
\end{IEEEkeywords}

\section{Introduction}

Electric vehicle (EV) adoption has surged recently, accompanied by significant investments in charging infrastructure to support widespread usage. One of the main actors in is infrastructure are the EV chargers, which are critical components in modern transportation infrastructure that enable seamless energy delivery to EVs. Therefore, EV chargers are potential targets for cyberattacks due to their interconnectedness on different information and communication technologies \cite{aldhanhani2024future}.
% \cite{qadir2024navigating, aldhanhani2024future}
Such attacks pose substantial risks, as they can lead to data breaches, service disruptions, physical damage to the charging infrastructure, or even harm to connected electric vehicles \cite{yuvaraj2024comprehensive}. With the growing reliance on these systems, developing effective and reliable methods for detecting malicious activities at EV chargers has become a critical research area \cite{acharya2020cybersecurity}.% \cite{ xuefeng2022risks}.

The cybersecurity of Electric Vehicle Supply Equipment (EVSE) has garnered significant attention, with recent works addressing various aspects of detecting and mitigating cyber threats \cite{ronanki2023electric}. In \cite{akbarian2024detection}, Akbarian et al. proposed a machine-learning-based framework for detecting cyberattacks aimed at manipulating EV charging prices. 
Girdhar et al. \cite{girdhar2021hidden} addressed the security of extreme fast charging stations by applying STRIDE threat modeling and developing a Hidden Markov Model to correlate anomalies in attack-defense scenarios.  Despite its strengths, this approach required detailed threat modeling and system-specific information, making it challenging to scale across diverse EV charging infrastructures.
Liu et al. \cite{liu2022false} investigated false data injection attacks on smart grid voltage regulation processes incorporating EVs. While effective, this work focused on grid-level attacks rather than localized EVSE-specific threats.
While existing works have advanced the state of EVSE cybersecurity, they  often rely on network-level data, detailed threat modeling, or pre-defined system behaviors, which may not be feasible for real-time applications. There remains a critical gap in developing a model-free, data-driven framework that leverages only locally available charger measurements, such as power consumption data, for attack detection.

\begin{figure}[b!]
\centering
\includegraphics[width=0.9\linewidth]{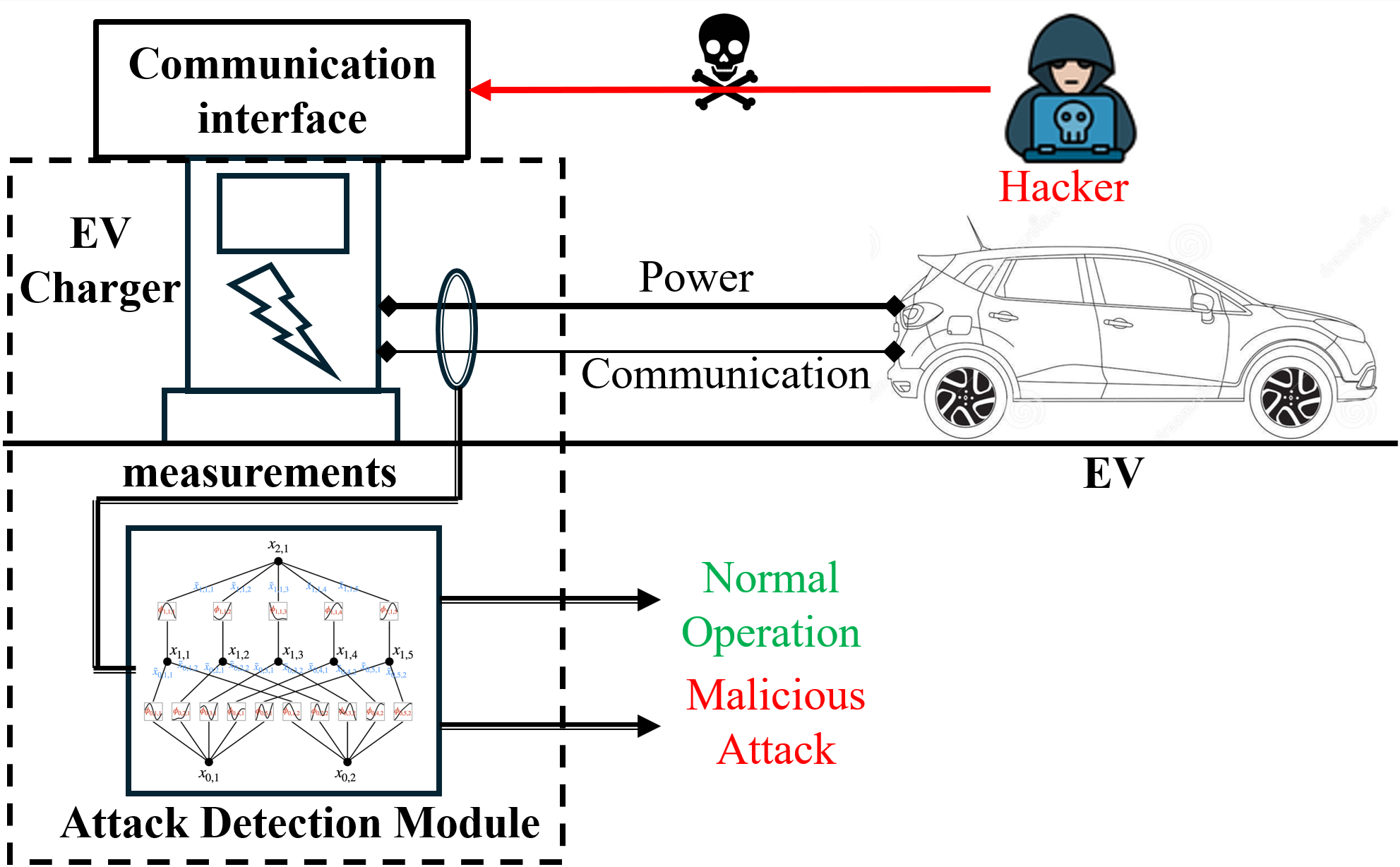}
\caption{The proposed KAN-based framework for detecting cyberattacks against EVSE.}
\label{fig:EVSE-KAN.jpg}
\end{figure}

Addressing this gap, this paper presents a Kolmogorov-Arnold Network (KAN)-based approach for detecting cyberattacks directly from EV charger measurements, providing an additional layer of security that is \textbf{accurate}, \textbf{interpretable}, lightweight, and suitable for deployment in diverse charging environments. The proposed approach is illustrated in Fig. \ref{fig:EVSE-KAN.jpg}.
By leveraging KANs, a novel class of neural networks introduced in 2024 based on the Kolmogorov-Arnold representation theorem \cite{liu2024kankolmogorovarnoldnetworks}, this approach effectively models nonlinear relationships in the charger's measurements.
% /
The KAN-based attack detection framework identifies abnormal patterns in charger measurements that may indicate cyberattacks, without requiring detailed network models. The choice of KAN is motivated by its capability in function approximation, particularly for high-dimensional, nonlinear data. Unlike traditional neural networks, KANs decompose complex multivariate relationships into sums of univariate functions, significantly enhancing interpretability and computational efficiency in the context of real-time cyberattack detection.
Finally, the proposed model is validated on a comprehensive set of scenarios, encompassing both malicious and normal charging conditions.

In the remainder of this paper, Section II depicts the problem formulation and threat model. Section III presents the proposed KAN-based approach to detect cyberattacks on electric vehicle chargers. In Section IV, the main performance evaluation results are discussed. Finally, Section V concludes the paper.

\section{Preliminaries and Threat Model}
 \label{section:Preliminaries}

\subsection{Electric Vehicle Charging Systems}

The EV ecosystem includes various components, including the EV itself, its battery management system (BMS), and the charging infrastructure, which encompasses the electric vehicle supply equipment (EVSE). These components interact dynamically to ensure the efficient transfer of energy from the grid to the vehicle, manage the storage of energy within the battery, and enable communication between the EV, charger, and grid.

Modeling electric vehicles (EVs), charging stations, and smart grids involves integrating various systems and technologies to create a coherent and efficient energy network. Here’s a detailed overview of how each component is modeled and their interactions. Electric vehicle modeling focuses on the vehicle's energy consumption, battery management, and operational behavior. Battery Voltage and SOC can be modeled using differential equations or empirical formulas such as
\begin{equation}
\scalebox{1}{
$V(t)=V_{oc} - IR - K(T) I$
}
\end{equation}
where $V(t)$ is the battery voltage at a time $t$, $V_{oc}$ is the open-circuit voltage, $I$ is the current, $R$ is the internal resistance, and $K(T)$ is the temperature-dependent resistance factor. Charging station modeling involves simulating the infrastructure for charging EVs, including power management and communication. It includes power supply, communication protocols, and control systems. Charging power can be expressed as $P = VI$, where $P$ is the charging power, $V$ is the voltage, and $I$ is the current. 
From a smart grid perspective, it is crucial to properly integrate EVs and their charging stations into the broader electrical grid, focusing on grid stability, energy management, and demand response.    
We model the EVs and charging infrastructure as a dynamic system with time-varying states, captured by a set of sensors. Let $S=\{s_1,s_2, \dots ,s_n\}$ represent the set of sensors, where $s_i(t)$ denotes the reading from sensor $i$ at time $t$. The system state is represented by the vector $X(t) = [s_1(t), s_2(t), \dots, s_n(t)]$. 
The cyberattack representation involves modeling and understanding potential cybersecurity threats and attacks that could compromise the vehicle's systems or data. In autonomous vehicles, cyberattacks can target communication networks, control systems, or data integrity. 
A malicious manipulation of the sensor data or communication signals can represent a cyberattack. The system state under attack at $t_a$ is given by $X_a(t_a) = X(t_a) + A(t)$, which represents the attack vector.

\subsection{Threat Model}

The threat model makes the following assumptions: the attacker has partial or full access to the EV charger’s communication network;   
stable power consumption patterns characterize the system’s normal operating conditions; and the detection framework has access only to power-related measurements, such as voltage, current, and derived power.
Therefore, the goal is to develop a detection framework to differentiate between normal charging scenarios and malicious cyberattacks against the charger, using only the EV charger's measurements. We focus in this paper on developing a cyberattack detection module that utilizes only these measurements as the only inputs. This is since these measurements are locally collected within the charger so they can be directly sent to the new attack detection model without being manipulated by the adversaries.
Consequently,  this problem can be formulated as a binary classification problem given a time-series input of power consumption measurements ${x}(t) = [x_1, x_2, \ldots, x_n]$, to classify each observation as either normal $(y = 0)$ or malicious $(y = 1)$, where $y$ is the attack detection model's prediction.

\section{A KAN-based Framework for Detection of Cyberattacks on EV Chargers}

Figure ~\ref{fig:EVSE-KAN.jpg} illustrates the proposed architecture for defending EV Supply Equipment against cyberattacks using the KAN. The architecture monitors power Measurements (including voltage, current, and derived power). In normal operation, the power consumption patterns adhere to predictable profiles, ensuring efficient and secure energy delivery to EVs. However, during malicious attacks, the power consumption patterns no longer adhere to known patterns. 
For this purpose, Kolmogorov-Arnold Networks (KANs) are used in this paper as tools to detect cyberattacks on EV  chargers.
The KAN leverages power measurements to differentiate between normal and malicious charging scenarios.
A KAN-based model can be implemented within each EV charger to detect cyberattacks. This model continuously processes the same measurements collected within the charger, such as power consumption measurements. Based on the charger's measurements, this model can be trained offline to differentiate between normal and cyberattack EV charging scenarios, which can be seen as a binary classification problem.
During the offline training, the network learns to recognize patterns associated with malicious activities and legitimate operations, using time-series data of power measurements as input. Once trained, the KAN model can be implemented directly within the EV charger’s system for online real-time monitoring of malicious behaviors that could indicate a cyberattack, e.g., unusual voltage or current patterns. This capability makes it an invaluable tool for maintaining the security of charging infrastructure, especially in smart grid environments.

KANs are based on the Kolmogorov-Arnold theorem, which states that any multivariate continuous function can be decomposed into a finite sum of univariate functions and addition operations. This property enables KAN to approximate complex, high-dimensional functions, capturing the intricate nonlinear patterns typical in power consumption data during cyberattacks.  In addition, unlike many deep learning models, KAN offers a more interpretable structure, which can be advantageous when analyzing the underlying causes of flagged events. This interpretability aids operators in understanding the nature of detected anomalies.
The KAN representation theorem provides the foundation for modeling. For a smooth multivariate function $f: [0, 1]^n \to \mathbb{R}$, the function can be expressed as

\begin{equation}
f(x) = \sum_{q=1}^{2n+1} \Phi_q \left( \sum_{p=1}^n \phi_{q,p}(x_p) \right)
\end{equation}

\noindent where  $\phi_{q,p}: [0,1] \to \mathbb{R}$  and  $\Phi_q: \mathbb{R} \to \mathbb{R}$ are  univariate functions. 
This formulation decomposes the multivariate problem into additive compositions of univariate functions. While the original Kolmogorov-Arnold representation is limited in expressiveness due to its shallow structure, we extend its capacity by introducing deeper architectures with arbitrary depths and widths.
In matrix form, the KAN representation is defined as

\begin{equation}
f(x) = {\bf \Phi}_{\text{out}} \circ {\bf \Phi}_{\text{in}} \circ {\bf x}
\end{equation}

\noindent where
\begin{equation}
{\bf \Phi}_{\text{in}} = 
\begin{pmatrix} 
\phi_{1,1}(\cdot) & \cdots & \phi_{1,n}(\cdot) \\ 
\vdots & & \vdots \\ 
\phi_{2n+1,1}(\cdot) & \cdots & \phi_{2n+1,n}(\cdot) 
\end{pmatrix}
\end{equation}

\noindent  and 

\begin{equation}
\quad
{\bf \Phi}_{\text{out}} = 
\begin{pmatrix} 
\Phi_1(\cdot) & \cdots & \Phi_{2n+1}(\cdot)
\end{pmatrix}
\end{equation}

\noindent By stacking multiple KAN layers, we generalize the representation to form a network. For a model with $L$ layers, the output is computed as

\begin{equation}
\text{KAN}({\bf x}) = {\bf \Phi}_{L-1} \circ \cdots \circ {\bf \Phi}_1 \circ {\bf \Phi}_0 \circ {\bf x}
\end{equation}

\begin{figure}[t!]
\centering
\includegraphics[width=0.8\columnwidth]{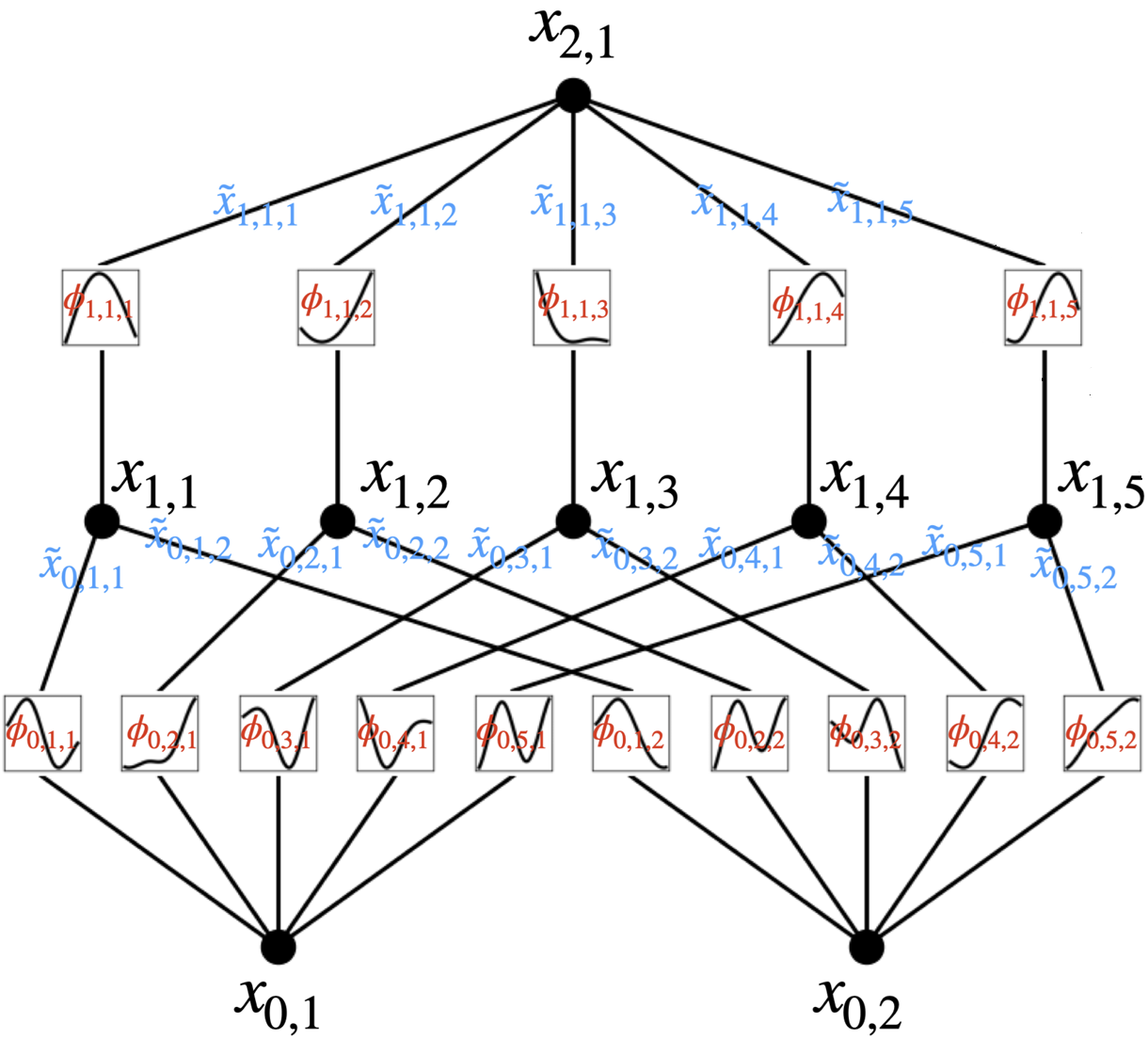}
\caption{High-level idea of KANs with activation functions on edges instead of nodes.}
\label{fig:KAN}
\end{figure}

\noindent Each KAN layer is a fully connected structure with 1D univariate functions on the edges, making the architecture interpretable. 
Therefore, a key difference between neural networks,  which have fixed activation functions on nodes and learnable weights on edges, and KANs is that KANs have learnable activation functions on edges and sum operations on nodes, as illustrated in Fig. \ref{fig:KAN}. 

The training process of a KAN model involves using labeled time-series data of power consumption measurements under normal and attack conditions. The KAN model learns the relationships between features and the corresponding labels, optimizing its parameters to maximize classification accuracy. Once trained, the model is lightweight and computationally efficient, making it suitable for deployment within the resource-constrained environments of EV chargers.

\section{Performance Evaluation Results}\label{section:results}

\subsection{Perfromance Evaluation Scenarios and Dataset}

\begin{figure}[t!]
\centering
\includegraphics[width=0.7\columnwidth]{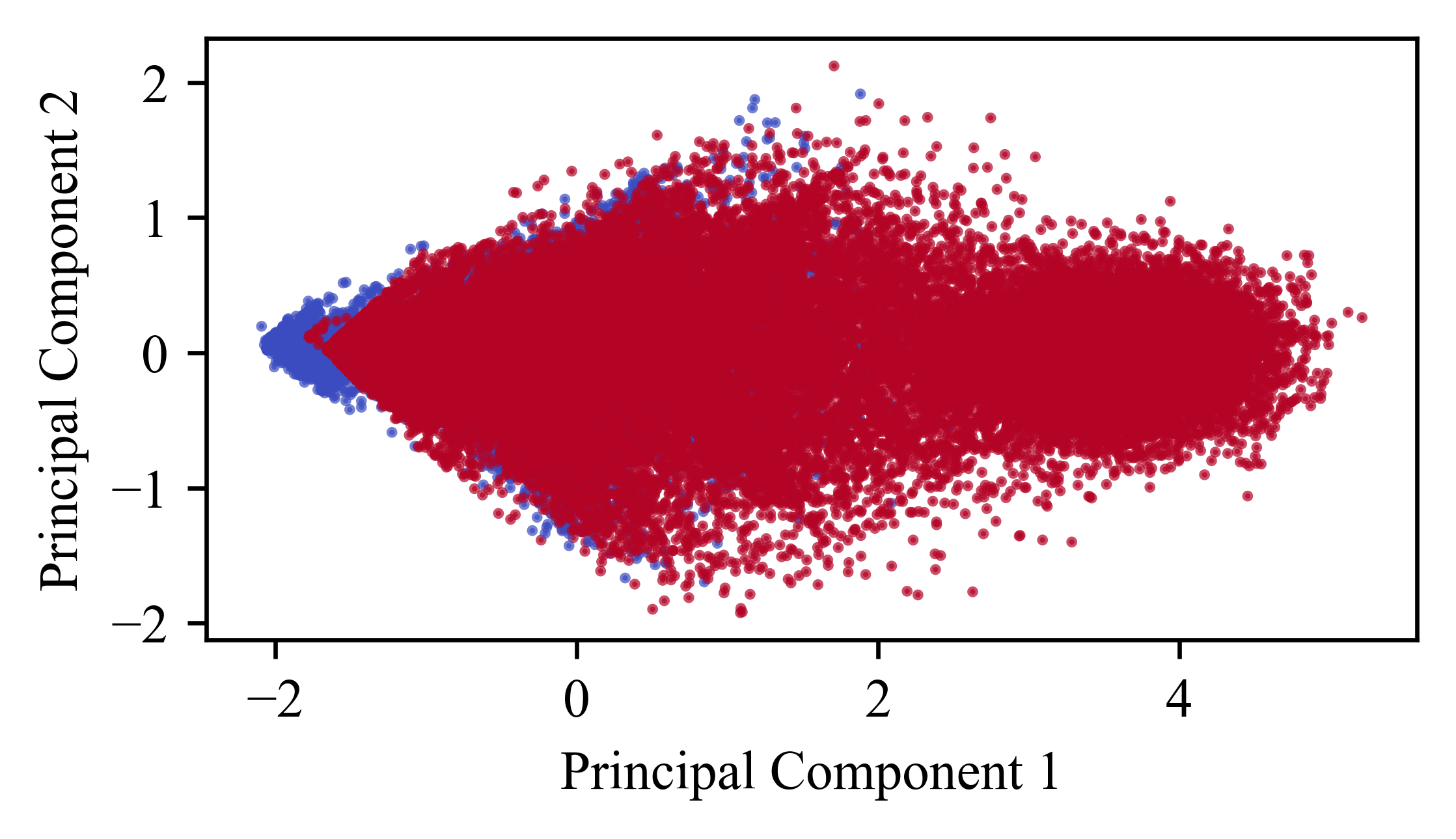}
\caption{Scatter plot of the dataset after applying PCA transformation, blue and red dots denote benign and malicious cases, respectively}
\label{fig:PCA_plt}
\end{figure}

To validate the proposed KAN approach, this study utilizes the dataset of EVSE locally-measured power consumption data under normal charging scenarios vs diverse cyberattacks, introduced in \cite{buedi2024enhancingthesis}. This dataset was generated using an experimental testbed incorporating real Electric Vehicle Supply Equipment (EVSE), a Raspberry Pi single-board computer, and standard industry communication protocols for EV charging infrastructure.
The dataset was collected in a controlled environment simulating real-world EV charging scenarios to analyze cyberattacks and their impact on EVSE systems. It includes diverse features such as the attacked charger’s voltage levels, current, and power consumption measurements under both normal charging conditions and various simulated cyber threats. Attack scenarios include backdoor intrusions, cryptojacking, SYN-flood, SYN-stealth, TCP-flood, and denial-of-service attacks, among others. These scenarios provide valuable insights into the effects of different attack vectors on the performance and security of EV charging systems, as detailed in \cite{buedi2024enhancingthesis}.
The dataset comprises a substantial number of samples, with 14,363 representing normal charging conditions and 100,935 capturing various cyberattacks on EV chargers. In each scenario, power consumption features, depicted in Table \ref{tab:features}, are recorded at a sampling rate of 1 sample per second. These features include:
Each sample in the dataset is properly labeled as either a normal scenario or an attack, making it suitable for training and validating machine learning models for cybersecurity analysis in EVSE systems.
Figure \ref{fig:PCA_plt} shows a scatter plot for the data after applying the Principal Component Analysis (PCA) to reduce the number of data dimensions down to 2. It
can be noticed that it is difficult to detect cyberattacks using only the power measurements, even after applying the PCA. Accordingly, in this paper, we aim to solve this problem by leveraging KANs due to their aforementioned merits including anticipated high accuracy in distinguishing between benign and malicious cases and enhanced interpretability.

\begin{table}[t!]
\centering
\caption{Utilized Features Based on the EV Charger Measurements}
\label{tab:features}
\begin{tabular}{ c|c |c }
\Xhline{3\arrayrulewidth}
\rule{0pt}{3ex} 
\textbf{Feature}  & \textbf{Description} & \textbf{ Unit} \\ 
\Xhline{2\arrayrulewidth}
\rule{0pt}{3ex} Feature 1 & Shunt voltage (voltage drop across the shunt & (V)
\\ 
  & resistor of the EVSE’s wattmeter) &
\\ 
\rule{0pt}{3ex} Feature 2 & DC bus voltage & (V) \\  
\rule{0pt}{3ex} Feature 3 & Current consumption of the EVSE & (A) \\  
\rule{0pt}{3ex} Feature 4 & Power consumption of the EVSE & (W) \\  
\Xhline{3\arrayrulewidth}
\end{tabular}
\end{table}

\subsection{Training of the KAN-based Cyberattack Detection Model  }

\begin{figure}[t!]
\centering
\includegraphics[width=.9\columnwidth]{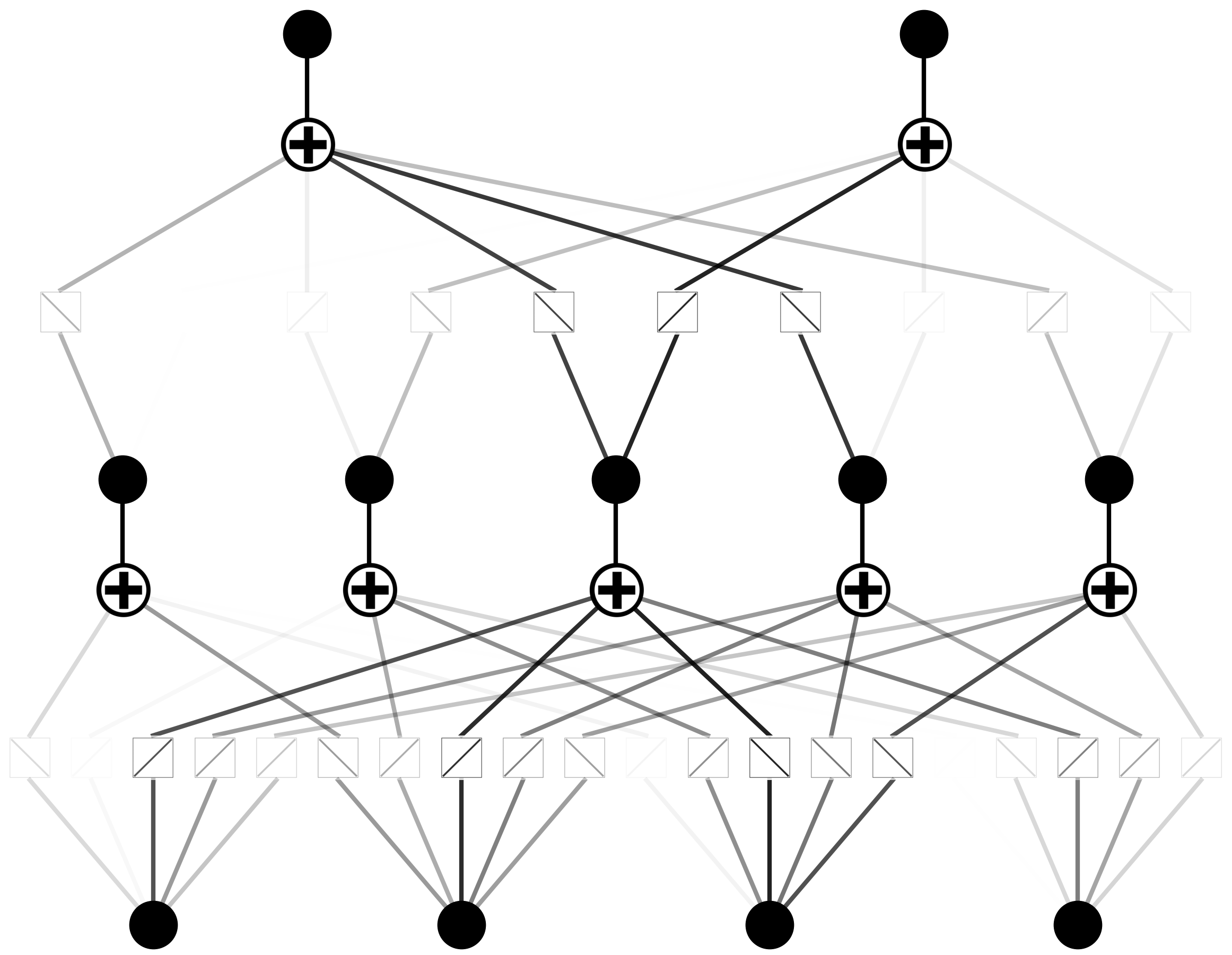}
\caption{An illustration of utilized KAN model's initial architecture with 5 hidden neurons, a cubic spline, and 3 grid intervals}
\label{fig:model_0_initial}
\end{figure}

The KAN model is trained and validated on 80\% of the dataset, with the remaining 20\% reserved for testing. The chosen configuration for the model is illustrated in Fig. \ref{fig:model_0_initial}.  In this illustration, which can be directly obtained after defining the KAN model, lighter-colored edges represent those with lower weights. Edges weights are randomly initialized at this stage. 
This model is trained on the 80\% of the dataset over 50 epochs. Cross entropy loss is used as the loss function, and to address class imbalance in the dataset, the model is optimized for balanced accuracy rather than average accuracy.  Afterward, the model’s performance is evaluated on the test set, comprising 20\% of the normal and cyberattack scenarios. 

\subsection{Testing Results}
\subsubsection{Accuracy of Trained KAN Model}

\begin{figure}[t!]
\centering
\includegraphics[width=0.3\columnwidth]{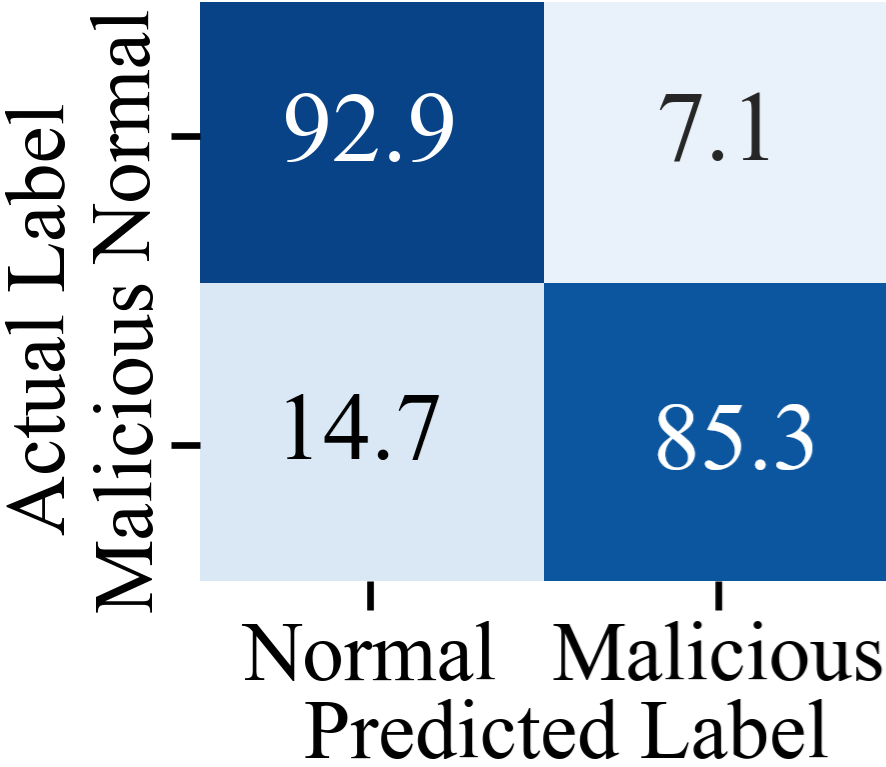}
\caption{Proposed KAN's confusion matrix in percentages.}
\label{fig:Conf_mtrx}
\end{figure}

The confusion matrix, shown in Figure \ref{fig:Conf_mtrx}, demonstrates promising results. This promising performance is shown in terms of the performance evaluation metrics--Precision, Recall and F1-Score \cite{saber2024unmasking}--depicted in Table \ref{tab:classification_report}, which are calculated from the viewpoint of the malicious cases.
The model achieves a true positive rate of 92.9\% for normal instances and successfully detects 85.3\% of cyberattack/malicious instances. However, a false positive rate of 7.1\% and a false negative rate of 14.7\% are observed, indicating that some false alarms occur, and a small portion of attacks remain undetected.
\Ablue{The observed false positives (7.1\% of
 normal instances misclassified as attacks) primarily result from model sensitivity to subtle anomalies in power measurements that resemble attack patterns.}
\begin{table}[t!]
\centering
\caption{Performance evaluation metrics results}
\label{tab:classification_report}
\begin{tabular}{c|c|c|c}
\Xhline{3\arrayrulewidth}
\rule{0pt}{3ex} 
\textbf{Metric} & \textbf{Precision} & \textbf{Recall} & \textbf{F1-score} \\  
\rule{0pt}{3ex} Value & 0.99 & 0.85 & 0.92 \\ 
\Xhline{3\arrayrulewidth}
\end{tabular}
\end{table}

Overall, the results are highly favorable. The model effectively detects most cyberattacks, significantly reducing the risk of undetected malicious activities. Nonetheless, false alarms suggest that some benign instances may be misclassified as attacks. This is a common challenge in systems designed for high security, where minimizing the risk of missed detections can sometimes lead to increased false positives. Further, the proposed KAN model outperforms other models, such as those reported in related works on the same dataset   \cite{buedi2024enhancingthesis}. This is depicted in Table \ref{tab:comparison}, where the average accuracy of the KAN model is compared to that of other baseline models, including Isolation Forest, Autoencoder, Local Outlier Factor, and One-Class SVM. In particular, the detection rate for cyberattacks is notably higher. The undetected attacks are primarily either stealthy ones designed to mimic normal behavior, or attacks that do not highly impact the power consumption of the EV charger, making them difficult to distinguish from normal instances based solely on the available measurements.

\begin{table}[t!]
\centering
\caption{Comparison 
% of average accuracy 
between KAN and baseline models}
\label{tab:comparison}
\begin{tabular}{ c|c}
\Xhline{3\arrayrulewidth}
\rule{0pt}{3ex} 
\textbf{Model} & \textbf{ Accuracy} \\ 
\Xhline{2\arrayrulewidth}
\rule{0pt}{3ex} Isolation Forest & 0.37 \\ \ 
\rule{0pt}{3ex} Autoencoder & 0.336 \\  
\rule{0pt}{3ex} Local Outlier Factor & 0.40 \\  
\rule{0pt}{3ex}  One-Class SVM & 0.41 \\  
% \rule{0pt}{3ex}  MLP & 0.56 \\  
\rule{0pt}{3ex} \textbf{Proposed KAN} &\textbf{ 0.89} \\ 
\Xhline{3\arrayrulewidth}
\end{tabular}
\end{table}

\subsubsection{Accuracy of KAN Model's Extracted Symbolic Formulas}

Further, as highlighted earlier, one key advantage of KANs is their enhanced interpretability. After training a KAN model, it is possible to derive a set of mathematical formulas that represent the model's decision-making process. 
That is, the predictions made by the trained KAN model, as a binary classifier, can be expressed through the following mathematical representation
\begin{equation}
\mathbb{1}\left\{ L_2 > L_1 \right\}
\end{equation}
% /
\noindent Here, \( \mathbb{1}\left\{ L_2 > L_1 \right\} \) is the indicator function, which outputs \textbf{1} if \( L_2 > L_1 \) (indicating a cyberattack) and \textbf{0} otherwise (indicating a normal scenario). The terms \( L_1 \) and \( L_2 \) represent the logits (outputs) of two formulas extracted from the trained KAN model. 
Each one of these formulas is a mathematical relationships between the KAN model's input features and the respective formula's   logit (output), offering a clear and interpretable representation of the model's decisions. This interpretability represents a notable advancement in deep learning-based cyberattack detection schemes. 
Accordingly, we obtained the mathematical formulas for the trained KAN model using automatic symbolic regression. 
Next, we tested the performance of these formulas combined as an attack detection tool. Our results indicate that the formulas achieve a testing accuracy of 87.28\%, i.e., on the testing dataset, compared to 87.68\% training accuracy on the training dataset, suggesting no overfitting. Even though this testing accuracy is still high, it is slightly lower than that of the full KAN model,  demonstrating a trade-off between interpretability and accuracy--an area for future research.

\subsection{Scalability and Computational Efficiency}
\Ablue{The proposed KAN model is designed for distributed deployment across individual EV chargers, eliminating centralized processing bottlenecks. Each EVSE independently processes its local power measurements, ensuring scalability across large-scale charging networks. To validate real-time feasibility, the model was tested on a PC with a 2.3 GHz CPU, a NVIDIA GeForce RTX 4070 GPU and 8 GB of dedicated memory, where the inference time is found to be 12.53 ms, i.e., per input sample. This lightweight performance, coupled with the model’s high accuracy, demonstrates its suitability for both edge and centralized computing environments.}

\section{Conclusion and Future Work}

This paper presented a novel explainable KAN-based approach to detect a variety of cyberattacks on EV chargers using only the charger's power consumption measurements. 
The proposed approach utilizes a KAN model, trained offline on both normal and malicious charging scenarios of the EV charger. Once trained, the model can be implemented within the charger to flag any malicious behavior that could indicate a cyberattack.
Our model has been validated on tens of thousands of normal and malicious charging scenarios, where it has shown superior accuracy compared to existing techniques. The proposed method not only enhances the security and reliability of EV operations but also offers an explainable solution, making it suitable for deployment in automotive networks where both security and interpretability are important.

\Ablue{Future researchers can
explore several key directions including: (1) minimizing the trade-off between the KAN's accuracy and interpretability; (2) integrating network-layer features to enhance detection accuracy; (3) investigating the model’s robustness to real-world conditions such as measurement errors and concurrent power quality issues during the EV charger's operation; and (4) optimizing the deployment of KANs for hybrid edge-cloud architectures to ensure scalable, low-latency operation across large EVSE networks. These efforts could extend KANs’ applicability to broader cybersecurity domains beyond EV charging systems.}

\ifCLASSOPTIONcaptionsoff
  \newpage
\fi

 \bibliographystyle{IEEEtran} 
 \bibliography{cas-refs.bib}

\end{document}